\documentclass[conference]{IEEEtran}
\IEEEoverridecommandlockouts
\usepackage{cite}
\usepackage{amsmath, amssymb, amsfonts}
\usepackage{algorithmic}
\usepackage{graphicx}
\usepackage{textcomp}
\usepackage{xcolor}
\usepackage{booktabs}
\usepackage{adjustbox}
\usepackage{subcaption}
\usepackage[capitalize]{cleveref}
\crefname{subsection}{Subsec.}{Subsecs.}
\crefname{section}{Sec.}{Secs.}
\Crefname{section}{Section}{Sections}
\Crefname{table}{Table}{Tables}
\crefname{table}{Tab.}{Tabs.}

\def\BibTeX{{\rm B\kern-.05em{\sc i\kern-.025em b}\kern-.08em T\kern-.1667em\lower.7ex\hbox{E}\kern-.125emX}}
\usepackage{textcomp}
\usepackage[absolute,showboxes]{textpos}
\TPMargin{4pt}

\newcommand{\copyrightstatement}{
    \begin{textblock}{0.84}(0.08,0.938)
         \noindent{\footnotesize{\copyright 2025 IEEE.
         Personal use of this material is permitted.
         Permission from IEEE must be obtained for all other uses, in any current or future media, including reprinting/republishing this material for advertising or promotional purposes, creating new collective works, for resale or redistribution to servers or lists, or reuse of any copyrighted component of this work in other works.

         \noindent
         Accepted for publication in Proceedings of the IEEE Intelligent Vehicles Symposium (IV),  Cluj-Napoca - Romania, 22-25 June 2025.}}
    \end{textblock}
}  

\begin{document}

  \title{Cross-Level Sensor Fusion with Object Lists via Transformer for 3D Object Detection}
  % \author{\IEEEauthorblockN{Anonymous Authors}}
  \author{\IEEEauthorblockN{1\textsuperscript{st} Xiangzhong Liu} \IEEEauthorblockA{\textit{fortiss GmbH} \\
  % \textit{fortiss GmbH}\\
  Munich, Germany \\ xliu@fortiss.org}
  \and \IEEEauthorblockN{2\textsuperscript{nd} Jiajie Zhang} \IEEEauthorblockA{\textit{Technical University of Munich} \\
  % \textit{Technical University of Munich}\\
  Munich, Germany \\ jiajie.zhang@tum.de}
  \and \IEEEauthorblockN{3\textsuperscript{rd} Hao Shen} \IEEEauthorblockA{\textit{fortiss GmbH} \\
  % \textit{fortiss GmbH}\\
  Munich, Germany \\ shen@fortiss.org}
  % \and \IEEEauthorblockN{4\textsuperscript{th} Alois C. Knoll} \IEEEauthorblockA{\textit{Technical University of Munich} \\
  % % \textit{Technical University of Munich}\\
  % Munich, Germany \\ knoll@in.tum.de}
  }
  \maketitle
  \copyrightstatement

  \begin{abstract}
    In automotive sensor fusion systems, smart sensors and Vehicle-to-Everything
    (V2X) modules are commonly utilized. Sensor data from these systems are typically
    available only as processed object lists rather than raw sensor data from traditional
    sensors. Instead of processing other raw data separately and then fusing
    them at the object level, we propose an end-to-end cross-level fusion concept with
    Transformer, which integrates highly abstract object list information with
    raw camera images for 3D object detection. Object lists are fed into a Transformer
    as denoising queries and propagated together with learnable queries through the
    latter feature aggregation process. Additionally, a deformable Gaussian mask,
    derived from the positional and size dimensional priors from the object
    lists, is explicitly integrated into the Transformer decoder. This directs
    attention toward the target area of interest and accelerates model training
    convergence. Furthermore, as there is no public dataset containing object
    lists as a standalone modality, we propose an approach to generate pseudo object
    lists from ground-truth bounding boxes by simulating state noise and false positives
    and negatives. As the first work to conduct cross-level fusion, our approach
    shows substantial performance improvements over the vision-based baseline on
    the nuScenes dataset. It demonstrates its generalization capability over
    diverse noise levels of simulated object lists and real detectors.
  \end{abstract}


  \section{Introduction}
  \label{sec:intro}

  \begin{figure}[t]
    \centering
    \includegraphics[width=1\linewidth]{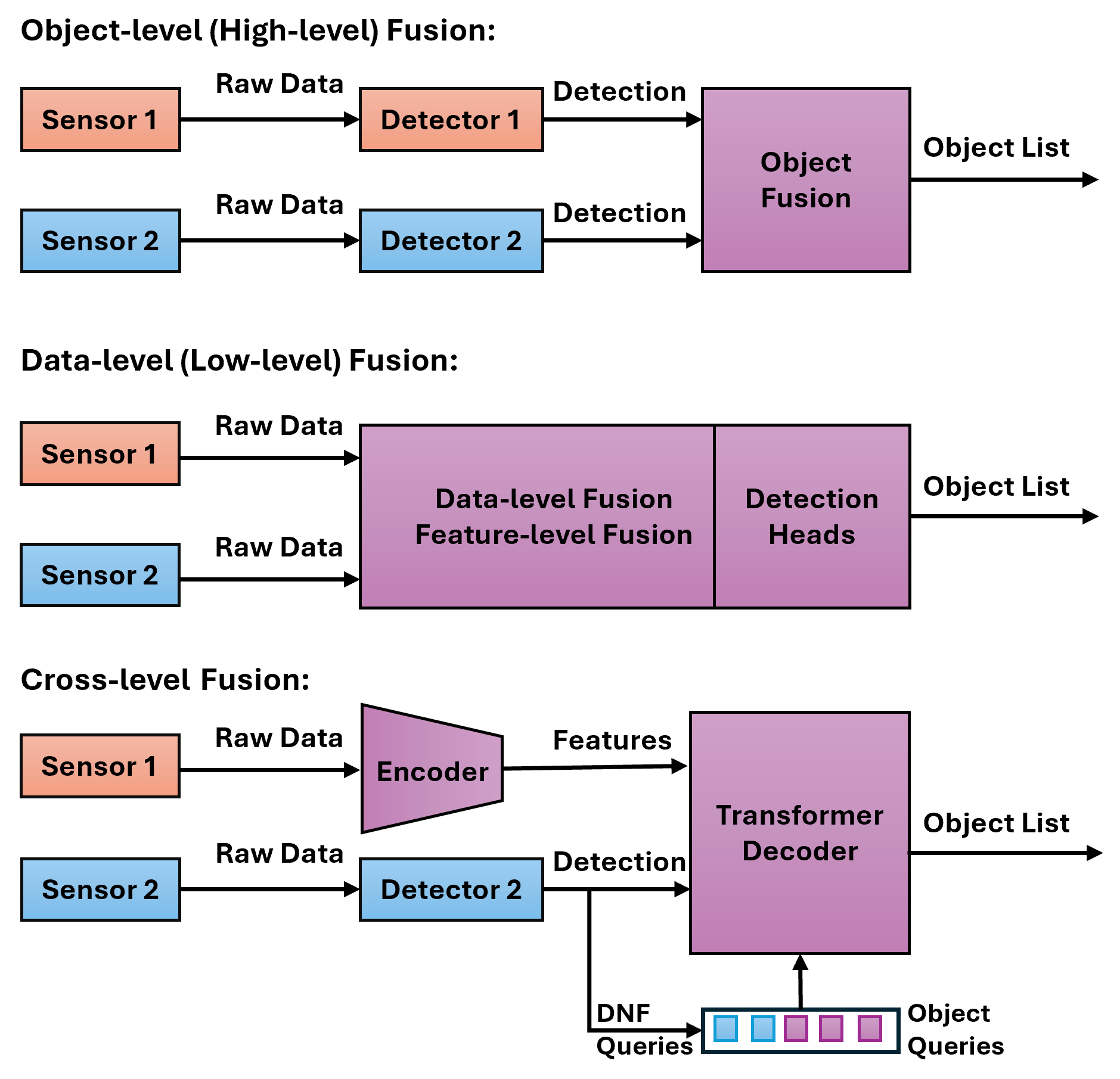}
    \caption{A brief illustration of 3 multi-sensor fusion strategies. In object-level
    fusion, distributed object detectors send the source object lists to a
    central fuser to generate a final set of detected objects. Conversely, in
    data-level fusion, data from multiple sensors are fed into a central fusion system
    to perform object detection. Cross-level fusion combines these approaches, comprehensively
    utilizing information from representations at different levels to enhance detection
    performance.}
    \label{fig:fusion_level}
  \end{figure}
  \vspace{-5pt}

  3D object detection is fundamental in autonomous driving and robotics. While vision-based
  methods have advanced, sensor fusion remains crucial for robust perception in
  dynamic scenarios by combining the strengths of multiple sensor modalities. Multi-sensor
  fusion systems are commonly categorized into three primary architectures based
  on different levels of abstraction: low-level, intermediate-level, and high-level
  fusion~\cite{castanedo2013fusion}, respectively fusing raw sensor signals,
  extracted features, and decisions or object lists. Low-level fusion
  theoretically offers optimal state estimation~\cite{thomopoulos1987optimal,chen2003limits,chong2011fusion},
  where recent advancements in deep neural networks (DNNs), particularly
  Transformers, have shown transformative capability.

  V2X technology is widely involved in collective perception for autonomous driving~\cite{ambrosin2019design},
  where the demand for data transmission efficiency has constrained the
  usability of low-level fusion. Furthermore, smart sensors~\cite{spencer2004smart}
  process data internally, such as MobileEye vision systems and Continental radars,
  and output only perceived object states other than raw data~\cite{matzka2008multifusion}.
  In these situations, fusion is only conducted at a high level~\cite{chong2000track}.
  No previous research has explored using end-to-end learning-based methods to fuse
  object lists with raw sensor data.

  Cross-level fusion refers to integrating high-level information to low-level feature
  fusion as illustrated in \cref{fig:fusion_level}. To achieve this, we extend DETR-like
  3D object detection frameworks by embedding object lists as DN queries.
  Moreover, we explicitly convert object lists' positional and dimensional
  information into an attention-weighting mask in the Transformer decoder. Treating
  the fusion process as an integral model rather than a separate step, as in
  high-level fusion, the framework can potentially create synergistic effects and
  achieve better efficiency and improved overall performance. High-level
  information offers context for interpreting raw sensor data, while raw data provides
  fine-grained feature details missing from the abstracted representations,
  reducing state estimation errors.

  We define object lists as structured sets of object detections containing
  position $(x,y,z)$, size $(l,w,h)$, orientation $(yaw)$, velocity $(v_{x},v_{y}
  )$, and class information with estimation errors. The cross-level fusion facilitates
  object detection by leveraging prior information from object lists.
  Consequently, we validate its performance using the nuScenes dataset~\cite{nuscenes2019}.
  Our approach significantly outperforms the vision-based baseline, revealing the
  substantial advantages of early integration of highly abstract information with
  raw sensor data.

  \begin{figure*}[t]
    \centering
    \includegraphics[width=2.0\columnwidth, keepaspectratio]{
      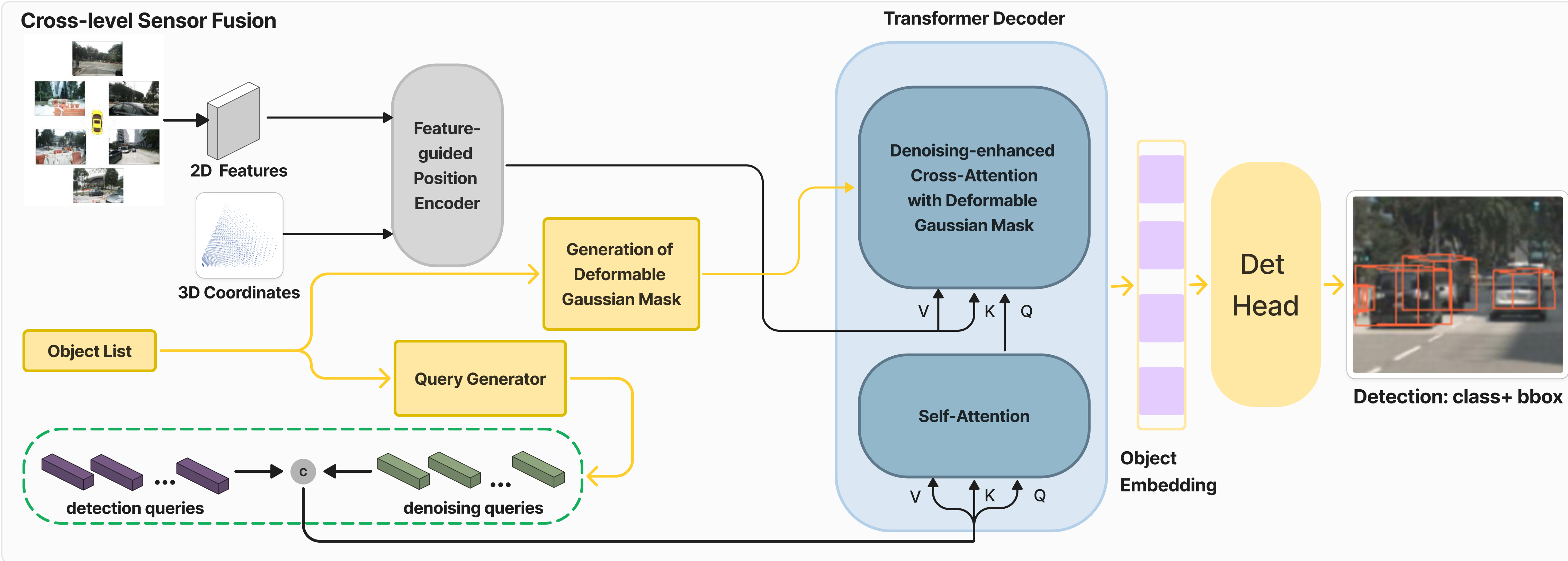
    }
    \caption{The illustration architecture of our overall cross-level fusion
    framework for 3D object detection. In particular, DN queries created from
    object lists are concatenated with detection queries from camera data to
    form denoising queries, which are fed into the self-attention module of the Transformer
    decoder for modality alignment. The information represented in the object
    lists is introduced into Transformer's cross-attention module using
    deformable multi-view multi-target Gaussian masks, which correlate with the 3D
    object representations from camera data through denoising processes. This
    approach facilitates more precise object positioning and classification. The
    object embeddings from the Transformer outputs are then processed by the detection
    head to generate object bounding boxes.}
    \label{fig:ourDQTrack}
  \end{figure*}

  In summary, the main contributions of this paper include:
  \begin{enumerate}
    \item We are the first to introduce the concept of cross-level sensor fusion
      with a Transformer-based framework for 3D object detection, combining data
      modalities of different abstraction levels in an end-to-end manner other
      than only at the high level.

    \item We leverage query denoising and explicit attention directing to implicitly
      and explicitly fuse object lists with raw image data. The approach
      integrates seamlessly into DETR-like 3D detectors as a plug-and-play
      module, requiring no additional parameters or computation, assuming a
      query-based architecture that supports attention-based conditioning.

    \item As no public dataset contains object lists as a standalone modality,
      we developed a pseudo object list generation (POLG) module that introduces
      diverse state noise, false positives and negatives to ground truth,
      simulating object detectors with varying performance levels.
  \end{enumerate}

  \section{Related Work}
  \label{sec:related_work}

  % \subsection{Sensor Fusion}
  % Sensor fusion technology in autonomous driving integrates data from multiple sensors to reduce uncertainty, merges their perception results into a cohesive representation, and therefore creates a more accurate environment model, enhancing autonomous vehicles' perception capabilities. Two common architectures used in multiple sensor fusion systems are data-level fusion and track-level fusion.
  \subsection{Object-level Fusion and Data-level Fusion}
  In object-level fusion (OF) distributed object detectors send the source
  object lists (object state estimation with error covariance) to a central
  fuser for more accurate decisions~\cite{castanedo2013fusion}, as shown in~\cref{fig:fusion_level}.
  OF often deploys explicit data association and Bayesian methods, like the Kalman
  filter with diverse motion models.

  OF enhances redundancy by incorporating multiple sensor sources, ensuring
  system reliability when facing individual sensor failure. Besides, it improves
  computational efficiency by reducing the data volume processed, as the fusion operates
  on structured object detections rather than raw data. Furthermore, object fusion
  facilitates modular scalability, allowing for the seamless integration of
  additional sensors into the system without significant architectural
  modifications~\cite{MathWorksFusion,matzka2008multifusion}. However, OF often
  faces data association errors due to limited object information, causing object
  state estimation inaccuracies.
  % Additionally, tracks from one sensor exhibit auto-correlation, and simply ignoring the correlation can lead to inconsistent fusion results~\cite{change1997optimaltrack}. The lack of access to (smart) sensors' internal states or variables prevents accurate track correlation modeling or de-correlation, ultimately resulting in degraded fusion accuracy or compromised estimation quality~\cite{matzka2008multifusion}.
  % \subsection{Data-level Fusion}

  In contrast, data-level fusion fully leverages detailed low-level data~\cite{MathWorksFusion}.
  Typically implemented using deep learning models, it follows three main
  approaches: early fusion, where raw sensor data is unified; late fusion, where
  data is processed separately before merging decisions; and intermediate fusion,
  which integrates features at an intermediate stage, balancing data interaction
  with computational efficiency~\cite{huang2022multi}. However, these methods are
  designed for dense and reach positional data like point clouds and are not directly
  applicable to object lists, which contain sparse, explicit information, such
  as bounding boxes. To effectively fuse object lists, alternative strategies
  are needed to harness their structured, concise nature.

  \subsection{Two-stage Detection and Proposal-level Fusion}
  The two-stage detection process typically involves two main steps: region proposal
  and region classification~\cite{girshick2014rich}. Initially, the system processes
  sensor data to produce feature maps, which are used to identify regions of
  interest (RoIs) by projecting 3D bounding boxes onto 2D feature maps. These methods
  use the object hypotheses generated in the initial detection stage to refine and
  validate object positions, enhancing the overall accuracy and robustness of
  the detection system. Notable works like MV3D~\cite{chen2017multi} and AVOD~\cite{ku2018joint}
  deploy multi-view aggregation for detection, while other methods incorporate
  Transformer decoders for advanced fusion. FUTR3D~\cite{chen2023futr3d} emphasizes
  a unified query structure called Modality-Agnostic Feature Sampler across different
  modalities and incorporates a Transformer decoder to avoid late fusion heuristics.
  TransFusion~\cite{bai2022transfusion} fuses LiDAR and image data, with the first
  decoder layer predicting bounding boxes from LiDAR and the second adaptively integrating
  image features. An image-guided query initialization strategy further enhances
  the detection of objects that are challenging to identify in point clouds. FusionRCNN~\cite{xu2023fusionrcnn}
  utilizes RoIPooling to obtain data samples within proposals and then employs intra-modality
  self-attention to enhance domain-specific features and a cross-attention mechanism
  to align both modalities. However, the region proposals and object queries,
  usually of a significant amount than the ground truth, still originate from raw
  data or features, which inherently contain substantially richer information
  and redundancy compared to the object lists.
  % \subsection{Denoising Mechanism for DETR-like Detectors}
  %  Relative approaches based on the DETR~\cite{detr} series mostly use the Hungarian matching algorithm to solve the bipartite graph matching problem. This involves finding an optimal matching pair between the predicted results and the true annotations. The discrete nature of the Hungarian matching algorithm and the stochastic nature of model training result in the matching of query to ground truth becoming a dynamic and unstable process. Especially in the early stage of model training, the targets can switch frequently, making learning the offsets of target positions very difficult. This instability leads to difficult and slow optimization, ultimately causing slow training convergence. To address the offset noise and uncertainty problems, DAB-DETR~\cite{liu2022dab} introduces an anchor-boxes-based and denoising mechanism, which greatly improves detection accuracy, especially in scenarios with complex target distributions. Building on this, DN-DETR~\cite{li2022dn} enhances the model's robustness by adding noisy object queries and introducing a denoising training strategy. This approach further reduces false positive and false negative and significantly accelerates the training convergence and inference of DETR models.

  \section{Methodology}
  \label{sec:methodology} The main aim is to implement a cross-level sensor
  fusion approach to fuse object list information with camera data end-to-end. For
  the overall architecture as shown in \cref{fig:ourDQTrack}, we retained
  the main structure of our baseline PETRv2~\cite{liu2023petrv2}, due to its query-based
  design and temporal modeling through a 3D position-aware attention mechanism.
  Images
  $I = \{I_{i}\in \mathbb{R}^{3 \times H_I \times W_I}, i = 1, 2, \dots, N\}$
  from N camera views are processed by a backbone network (e.g., VoVNetCP~\cite{Lee_2020_CVPR})
  to extract 2D multi-view features
  $F^{2d}= \{F_{i}^{2d}\in \mathbb{R}^{C \times H_F \times W_F}, i = 1, 2, \dots,
  N\}$. 3D coordinate generator discretizes the camera frustum space into a 3D meshgrid,
  transforming these coordinates into the 3D world space using camera parameters.
  2D features and 3D coordinates from neighboring frames are concatenated and processed
  by the 3D position encoder (PE) to generate 3D position-aware features. These
  features serve as keys and values in the Transformer decoder.

  We exploit object-centric query denoising (QDN) with object lists by combining
  denoising and learnable queries, which interact further with feature and positional
  encoding to extract object embeddings for detection. QDN reduces state
  uncertainty in the original object list and enriches DN queries by
  incorporating image features. Explicit attention directing utilizes positional
  and spatial information of the object list to create soft connections between the
  object priors and multi-view image features during cross-attention.

  To simulate realistic black-box detectors, we generate pseudo object lists
  with stochastic variations by sampling the ground truth bounding boxes (position,
  size, orientation, velocity, and labels). We apply configurable noise,
  including Gaussian position/size noise, object drops (detection failures),
  false positives (spurious detection), and label misclassification. Noise
  levels vary randomly within set limits to better mimic estimation errors and
  detection inaccuracies. However, real-world object lists could introduce
  additional uncertainty, which may affect fusion performance. Addressing this
  requires to selectively trust object lists based on confidence scores or
  consistency across frames.

  \subsection{Object-centric Query Denoising}
  \label{subsec:DNF}

  Object-centric query denoising is firstly introduced in DN-DETR~\cite{li2022dn}
  to tackle the issues with the instability of bipartite graph matching and inconsistent
  optimization goals during early training stages of DETR. This method
  initializes queries with noisy ground-truth bounding boxes, termed ``denoising
  queries'', aiming to reconstruct the original, clean boxes. The mechanism of
  DN-DETR series aligns with our assumption for cross-level fusion of object
  lists with raw sensor data using Transformers. The object list can be seen as noisy
  samplings of the ground truth with with estimation errors, FPs, and FNs, which
  can be directly converted to DN queries for DETR-like detectors. Besides, the
  built-in data processing systems to produce the object lists are black box, providing
  no information to de-correlate the object lists for optimal fusion~\cite{matzka2008multifusion}.
  However, learning-based approach enable end-to-end training and optimization,
  without the need for manual feature engineering or understanding of the
  underlying data processing mechanisms. Denoising enables the model to learn more
  robust feature representations that are less sensitive to variations and perturbations
  and to focus its attention more effectively for more precise object positioning
  and classification. Meanwhile, the queries sourced from the object list are enriched
  with raw data feature during cross-attention, and prepared for concatenation
  with extracted object embedding to the data association module, while DN-DETR discards
  denoising queries during inference. Introducing multiple denoising groups
  provides diverse training scenarios with a larger variety of noise levels, which
  accelerates the convergence and also prevents the overfitting to the GT-like data.

  Moreover, DN-DETR highlights that feeding ground truth-like information
  directly into the Transformer during training can cause over-reliance, degrading
  performance during inference. Hence, an extra attention mask regulating the visibility
  of two parts of queries is incorporated into the self-attention mechanism in the
  decoder to prevent information leakage~\cite{li2022dn}. In QDN, we discard the
  attention mask which prevents the self attention so that the detection queries
  can get useful information from the denoising part. The co-occurrence between
  raw data and object list facilitate the prediction of objects on raw data.
  % This approach addresses the limitations of both information sources. High-level information provides valuable contextual data, aiding in the interpretation of raw sensor data and accelerating the learning process. I.e. explicit positional and dimensional priors from object lists enhance spatial awareness and speed up model convergence. Conversely, raw data from camera images offer fine-grained feature details that high-level representations may miss or discard during distributed processing, helping to reduce state estimation errors during learning.
  \subsection{Explicit Attention Directing}
  \label{subsec:object_list_prior}

  \begin{figure}[t]
    \centering
    \includegraphics[width=1\linewidth]{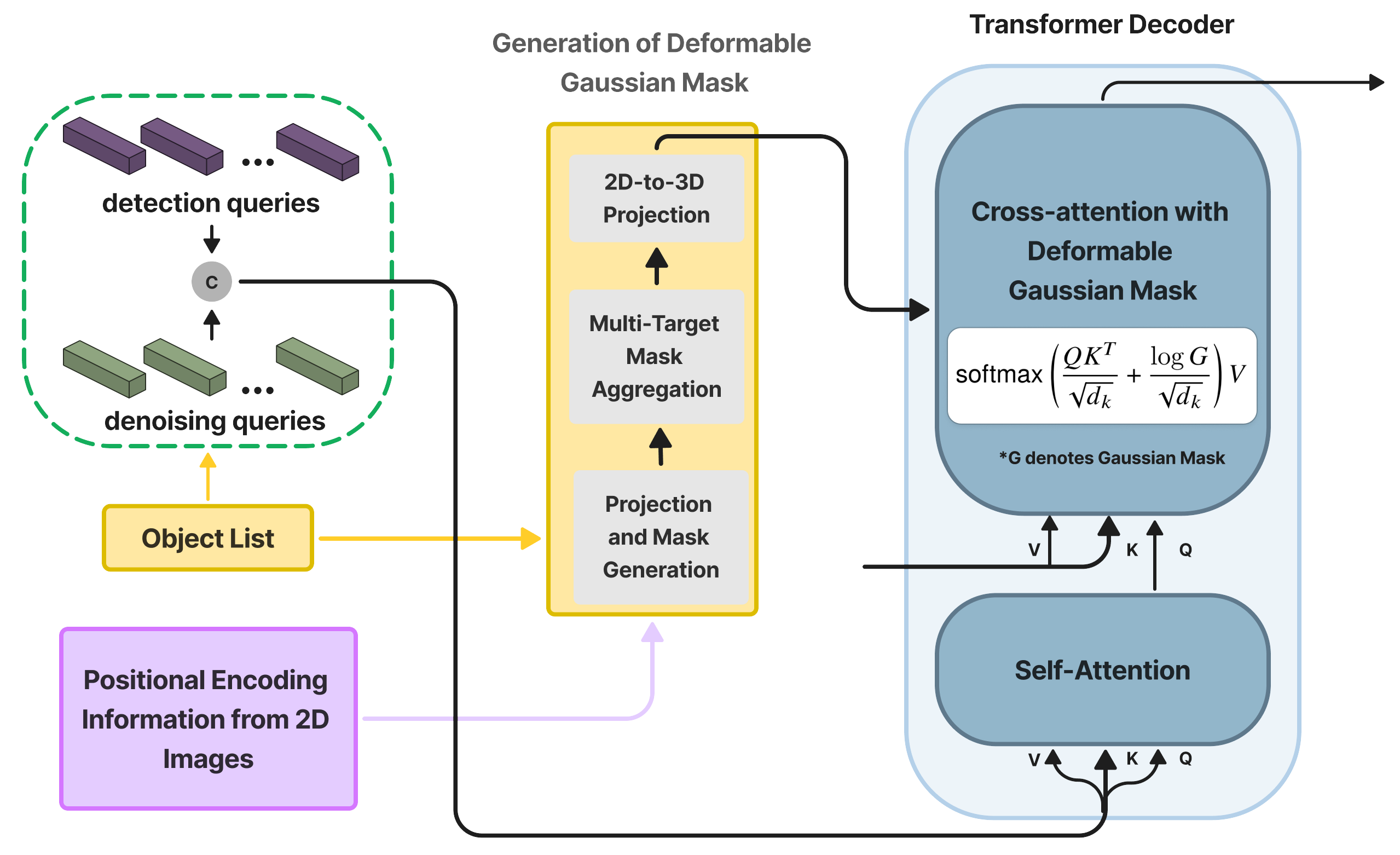}
    \caption{The structure of the deformable multi-view, multi-target shared Gaussian
    mask generation and its implementation in the Transformer decoder. Object
    list information is combined with positional encoding information from 2D
    images captured by the camera. Our SMCA approach generates Gaussian masks and
    transits these reshaped Gaussian masks to the cross-attention module,
    refining the attention mechanism by reweighting the attention maps.}
    \label{fig:gaussianmask}
  \end{figure}
  In addition to implicitly learning through the attention mechanism, we utilize
  a deformable Gaussian mask to adjust positional attention during cross attention,
  inspired by Spatially Modulated Co-Attention (SMCA)~\cite{gao2021fast}. SMCA creates
  a soft link between the object list and the raw image data to dynamically emphasize
  relative position during the learning process.

  While the SMCA approaches referenced in ~\cite{bai2022transfusion,gao2021fast}
  establish a one-to-one alignment between queries and targets constrained by a
  Gaussian mask around the initial query, PETRv2 generates a large amount of random
  queries without predefined links to the Gaussian circles around targets. To
  solve this issue, we implement a one-to-many assignment of detection queries to
  targets within the Gaussian mask. This allows queries to map to any target or
  even a no-object class after training.

  ~\cref{fig:gaussianmask} provides a detailed flowchart of the process to compute
  and integrate the Gaussian mask into the Transformer decoder. Initially, object
  position data from the object lists and position encoding of objects in the 2D
  images are ported to the mask generator. With the projection matrices of
  cameras, we project their corresponding positions in multi-view images,
  subsequently resulting in a 2D Gaussian weighting mask $M$ for each target. The
  mask generation process resembles the one applied in ~\cite{zhou2019objects}:
  \begin{equation}
    M_{ij}= \exp \left( -\frac{(i-cx)^{2}+(j-cy)^{2}}{\sigma*r^{2}}\right),
  \end{equation}
  where $(i, j)$ representing the mask’s spatial indices, $(cx, cy)$ as the 2D
  projected center, $r$ as the bounding box radius, and $\sigma$ as a parameter
  controlling Gaussian distribution bandwidth.

  % \begin{figure}[t]
  %   \centering
  %   \includegraphics[width=1\linewidth]{img/gaussian.PNG}
  %   \caption{Visualization of target Gaussian circles in our Gaussian mask maps.
  %   The Gaussian masks are created from the object lists for the foreground.
  %   Some ground truth targets are randomly dropped out during the generation
  %   process of object lists, such as the left white car in the button left image.}
  %   \label{fig:gauss_vis}
  % \end{figure}
  The individual masks for each target within an image are combined to form a
  multi-target shared Gaussian mask. The merge mask is then reshaped and transmitted
  to the 3D space. The attention reweighting between queries and feature maps
  involves element-wise addition of Gaussian weights to the cross-attention map
  across all attention heads. Consequently, the deformable Gaussian masks refine
  the attention mechanism, allowing each detection query to precisely concentrate
  on image areas that include projected targets from the object list.

  For the modified attention reweighting mechanism, we re-formulate the
  corresponding expression:
  \begin{equation}
    \text{Attention}(Q,K,V) = \text{softmax}\left(\frac{QK^{T}}{\sqrt{d_{k}}}+\frac{\log{G}}{\sqrt{d_{k}}}
    \right)V,
  \end{equation}
  where $G$ denotes our Gaussian masks. we incorporate a logarithmic function to
  prevent extreme values in Gaussian mask from overpowering the overall
  attention score without introducing negatives to attention map. This helps to control
  the attention scores of specific positions, ensuring they do not become
  excessively large and dominate the overall prediction. Additionally, we retain
  the operation of dividing by $\sqrt{d_{k}}$ from the original attention score
  calculation to improve model stability and training efficiency.
  % \cref{fig:gauss_vis}
  % visualizes the Gaussian circles on our Gaussian mask, which are generated by processing
  % the object lists by our SMCA approach.

  \begin{table*}
    [t!]
    \centering
    \begin{adjustbox}
      {max width=0.9\textwidth}
      \begin{tabular}{l|c|ccccccc}
        \toprule                              % \textbf{Tracking} & \textbf{Backbone} & \textbf{AMOTA} $\uparrow$ & \textbf{AMOTP} $\downarrow$ & \textbf{MOTA} $\uparrow$ & \textbf{MOTP} $\downarrow$ & \textbf{RECALL} $\uparrow$ & \textbf{IDS} $\downarrow$ \\
        % \midrule
        % DQTrack~\cite{li2023end} & V2-99 & 0.4395 & 1.2859 & 0.3812 & 0.7966 & 0.5564 & \textbf{1445} \\
        % SMCA-DNF-DQTrack & V2-99 & \textbf{0.5864} & \textbf{1.0991} & \textbf{0.5312} & \textbf{0.7595} & \textbf{0.6725} & 2040 \\
        % \midrule
        \textbf{Detection}                   & \textbf{Backbone} & \textbf{NDS} $\uparrow$ & \textbf{mAP} $\uparrow$ & \textbf{mATE} $\downarrow$ & \textbf{mASE} $\downarrow$ & \textbf{mAOE} $\downarrow$ & \textbf{mAVE} $\downarrow$ & \textbf{mAAE} $\downarrow$ \\
        \midrule PETRv2~\cite{liu2023petrv2} & V2-99             & 0.4038                  & 0.4136                  & 0.7208                     & 0.7094                     & 1.5492                     & \textbf{0.4084}            & 0.1914                     \\
        our model                            & V2-99             & \textbf{0.5685}         & \textbf{0.5499}         & \textbf{0.6334}            & \textbf{0.2558}            & \textbf{0.5006}            & 0.4860                     & \textbf{0.1885}            \\
        \bottomrule
      \end{tabular}
    \end{adjustbox}
    \caption{Comparison of cross-level fusion with baselines on the nuScenes val
    set. Image resolution of $800\times320$ as the inputs.}
    \label{table:over_baseline}
  \end{table*}

  \section{Experiments}
  \label{sec:Experiments}

  \subsection{Experimental Setup}
  \label{subsec:Experimental_Setup}

  \begin{table}[t]
    \centering
    \begin{adjustbox}
      {max width=0.9\textwidth}
      \begin{tabular}{c|cc|cc}
        \toprule \textbf{}      & \multicolumn{2}{c|}{\textbf{w Object Lists}} & \multicolumn{2}{c}{\textbf{w/o Object Lists}} \\
        \midrule \textbf{Model} & \textbf{mAP$\uparrow$}                       & \textbf{NDS$\uparrow$}                       & \textbf{mAP$\uparrow$} & \textbf{NDS$\uparrow$} \\
        \midrule DN-PETRv2      & 0.139                                        & 0.324                                        & \textbf{0.419}         & \textbf{0.507}         \\
        QDN-PETRv2              & 0.491                                        & 0.526                                        & 0.287                  & 0.404                  \\
        SMCA-QDN-PETRv2         & \textbf{0.550}                               & \textbf{0.569}                               & 0.299                  & 0.416                  \\
        \bottomrule
      \end{tabular}
    \end{adjustbox}
    \caption{Robustness of the cross-level fusion framework against loss of the
    modality of object lists. DN, QDN and SMCA denotes respectively original
    Denoising mechanismus, Query DeNoising, and Spatially Modulated Co-Attention.}
    \label{tab:robustness_analysis}
  \end{table}
  \paragraph{Dataset}
  We chose the nuScenes dataset~\cite{nuscenes2019} as our experimental data source,
  which is a popular benchmark widely used for 3D object detection and multi-object
  tracking tasks in large-scale autonomous driving. The nuScenes dataset provides
  synchronized data from 6 cameras surrounding the ego car, capturing 10 object
  categories for detection and 7 moving classes for tracking in a 360\textdegree-field
  of view at a frequency of 12Hz. It contains 700, 150 and 150 scenes respectively
  in the training, validation and test sets.

  \paragraph{Data Preparation}
  Our model and data processing pipeline are based on the MMDetection3D~\cite{mmdet3d2020}
  framework, facilitating efficient code development. Besides the multi-view images,
  the 3D annotations are imported into our training/testing pipeline with
  MMDetection3D to generate pseudo-object lists. We manually adopted a unified setup
  for all training experiments regarding the object list generation with max noise
  standard deviation ratio of 0.06, max drop rate of 0.2, max false positive rate
  of 0.1, and max label change rate of 0.3.

  \paragraph{Implementation Details}
  Our framework adopted detection baseline PETRv2 with modified denoising implementation
  and SMCA. In particular, we utilize 6 Nvidia RTX A5000 GPUs with 24GB of RAM in
  our experiments. The model is trained with a batch size of 1 due to the memory
  limit of the GPU for 24 epochs. we employ the AdamW optimizer with an initial learning
  rate of $2.0\times10^{-4}$. By default, the number of queries and threshold parameters
  for detection stay the same as baseline.

  % \paragraph{Evaluation Metrics} We evaluated the performance on the test subset of the nuScenes dataset and considered the following evaluation metrics, which are widely used for the open-loop evaluation on nuScenes: mean average precision (mAP) and nuScenes detection score (NDS) for the 3D object detection task, as well as average multi-object tracking accuracy (AMOTA) and average
  % multi-object tracking precision (AMOTP) for the 3D object tracking task.

  \subsection{Experimental Results}
  \subsubsection{Comparison to Baseline}
  \cref{table:over_baseline} exhibits the detection performance of our cross-level
  fusion framework over the vision-based baseline. Our model shows substantial improvements
  over the baseline models in most key performance metrics with the trade-offs of
  a decrease in velocity estimation accuracy (mAVE). The degraded velocity
  estimation may originate from the lack of velocity denoising, when the DN-queries
  only focus on position and size. A potential solution is to extend query
  initialization to incorporate velocity perturbation and refinement, which could
  improve motion estimation while maintaining strong detection accuracy. These
  results suggest that SMCA and QDN enhancements are generally effective and
  incorporating high-level information early in the feature level can
  significantly improve overall performance of the object detection with raw
  sensor data.

  \subsubsection{Ablation Study}
  % We validate the overall model performance improvement by integrating different modules designed in \cref{sec:methodology} step by step through ablation experiments. These include the introduction of a priori object queries into the learning process, concatenated with detection queries from camera data as denoising queries (DNF), and the incorporation of deformable Gaussian masks generated by the SMCA approach into the cross-attention mechanism to construct soft correlations between the input queries (SMCA-DNF). \cref{tab:ablation_study} shows the results of our ablation experiments.
  % From the results, both mechanisms contribute to the improvement. DNF brings significant benefits in tracking tasks compared to detection, which aligns with our assumption that DNF enhances tracking efficiency. While we observe that SMCA yields stronger improvements in detection. We can infer that the tracking improvements from SMCA primarily stem from the better detection results.

  Here we explore the effect of two key components in our design: query denoising
  (QDN) and explicit prior integration in attention mechanism with SMCA. As
  shown in~\cref{tab:ablation_study}, with QDN, PETRv2 already improves the
  detection performance by 8.01\% NDS and 2.30\% mAP. With SMCA, the performance
  is further improved by 5.94\% NDS and 4.30\% mAP, which shows a large margin compared
  to the baseline. The results confirm the synergistic benefits of combining
  implicit cooperation of object list information via QDN and explicit prior
  integration with SMCA in our framework. QDN improves detection performance,
  supporting our assumption that it enhances detection efficiency. While SMCA also
  brings significant benefits, its impact primarily comes from better feature
  alignment and attention guidance. Both modules add minimal overhead, as QDN
  replace part of the original queries without increasing total amount and
  Gaussian mask generation is efficiently implemented and scales linearly with
  the typically small object number($<50$).

  \begin{table}[t]
    \centering
    \begin{adjustbox}
      {max width=0.5\textwidth}
      \begin{tabular}{l|cc|cc}
        \toprule \textbf{}      & \multicolumn{2}{c|}{\textbf{Modules}} & \multicolumn{2}{c|}{\textbf{Detection Metrics}} \\
        \midrule \textbf{Model} & QDN                                   & SMCA                                           & NDS$\uparrow$   & mAP$\uparrow$   \\
        \midrule PETRv2         &                                       &                                                & 0.4104          & 0.5025          \\
        QDN                     & $\checkmark$                          &                                                & 0.4905          & 0.5255          \\
        SMCA-QDN                & $\checkmark$                          & $\checkmark$                                   & \textbf{0.5499} & \textbf{0.5685} \\
        \bottomrule
      \end{tabular}
    \end{adjustbox}
    \caption{The results of a step-by-step ablation study to evaluate the
    performance improvements of our proposed model over the baseline. The study involved
    progressively introducing different components of our framework.}
    \label{tab:ablation_study}
  \end{table}
  \subsubsection{Robustness against Modality Loss }
  From an engineering perspective, when raw data is unavailable, the object list
  should be directly forwarded to the decision-making module. Therefore, we define
  modality loss here as the absence of object lists. QDN differs from DN by
  removing the attention mask regulating the visibility between DN-queries and
  randomly-initialized queries. The significant performance drop in detection, a
  decrease by 25\% in mAP on our model under the absence of object lists, shows
  the reliability of Transformer decoder on the DN-queries during inference. Interestingly,
  the explicit prior integration via SMCA can slightly mitigate the degeneration.
  However, this strong sensitivity to presence of object lists is worthy of
  further exploration. A potential solution is to introduce masked-modality training
  as in CMT~\cite{yan2023cross}, where object lists are randomly masked during
  training. While we have not yet implemented this, future work could
  investigate its effectiveness in improving robustness to missing object lists.
  Additionally, we investigate the direct application of the original denoising
  mechanisms for the fusion of object lists. However, the results were
  unexpectedly suboptimal. When incorporating auxiliary high-level modal information
  into DN-queries during inference, the performance degradation was even more
  pronounced than with QDN without object lists. When injecting denoised queries
  during inference, the learnable queries may deviate from their intended
  targets.

  \subsubsection{Generalization with Real and Synthetic Object Lists}
  % The findings in \cref{tab:noise_levels} demonstrate our model's generalization
  % capacity across varying object list sources. With synthetic lists, performance
  % correlates directly with noise levels, but the model maintains robustness under
  % moderate noise conditions. To validate our approach with realistic inputs, we conducted
  % experiments using object lists from actual 3D detectors. CenterPoint~\cite{yin2021center}
  % produced object lists yielding fusion performance (mAP: 0.469, NDS: 0.531) comparable
  % to moderate-noise synthetic data. PointPillars~\cite{lang2019pointpillars}, with
  % lower baseline performance (mAP: 0.343, NDS: 0.491), resulted in proportionally
  % reduced but still significant improvements (mAP: 0.380, NDS: 0.470) over the
  % vision-only baseline. These results confirm both the robustness of our CLF framework
  % and the fidelity of our POLG noise settings.
  The findings in \cref{tab:noise_levels} demonstrate our model's generalization
  capacity across different object list sources. With synthetic object lists, performance
  correlates directly with noise levels, while maintaining robustness under
  moderate noise conditions. To validate our approach with realistic detections,
  we conducted experiments using object lists from actual 3D detectors. With CenterPoint~\cite{yin2021center}
  ( mAP 0.564, NDS 0.652), CLF achieves performance (mAP: 0.469, NDS: 0.531) comparable
  to moderate-noise synthetic data. PointPillars~\cite{lang2019pointpillars}, with
  lower performance (mAP: 0.343, NDS: 0.491), still provides significant improvements
  on NDS over the vision-only baseline. These results confirm both the
  effectiveness of our cross-level fusion approach with real detector outputs
  and the fidelity of our POLG noise parameters as reasonable approximations of
  actual detector behavior.
  \begin{table}
    \centering
    {\small{ \begin{tabular}{c c c}\toprule Object List Source & mAP$\uparrow$ & NDS$\uparrow$ \\ \midrule POLG (0.03, 0.1, 0.05, 0.1) & 0.590 & 0.593\\ \textbf{POLG (0.06, 0.2, 0.1, 0.3)} & 0.550 & 0.569 \\ POLG (0.1, 0.3, 0.2, 0.5) & 0.487 & 0.530 \\ \midrule CenterPoint (0.564/0.652) & 0.469 & 0.531 \\ PointPillars (0.343/0.491) & 0.387 & 0.478 \\ \bottomrule\end{tabular} }}
    \caption{Generalization capability with different object list sources. For PLOG
    lists, the parameters follow (std, drop, FP, label noise) format. Bold
    indicates settings used in training. For real detector, the mAP/NDS scores
    are reported.}
    \label{tab:noise_levels}
  \end{table}

  \section{Conclusion}

  % In this work, we propose a novel 3D object tracking framework that integrates explicit prior object list information with raw sensor data in an end-to-end manner, significantly enhancing scene perception and performance in object detection and tracking. Our key contributions include: (1) the introduction of a comprehensive end-to-end framework utilizing cross-level sensor fusion, surpassing traditional models in the nuScenes dataset evaluation; (2) the innovative DNF concept, which fuses high-level object lists with raw sensor data to improve accuracy and training efficiency; and (3) a pseudo object list creation approach to generate customized experimental data from open-source datasets. Future work could involve incorporating additional motion state information to further refine the model's performance.
  % In summary, our method SMCA-DNF demonstrates substantial improvements over baseline models in both tracking and detection tasks on the nuScenes validation set. These results validate the effectiveness of integrating highly abstract object list information with raw camera images, along with the use of deformable Gaussian masks in the Transformer decoder. The robustness study further supports the advantage of our approach, especially when object tracks are incorporated. The ablation study highlights the individual and combined contributions of DNF and SMCA to the overall performance enhancements. Future work will focus on optimizing identity consistency in tracking and enhancing velocity estimation in detection.

  Object lists contain high-level object information typically fused at the
  decision level. Our cross-level fusion approach introduces this high-level information
  into the low-level feature extraction process for 3D object detection. The
  denoising mechanism enhances the model's ability to learn resilient features, thereby
  improving positional accuracy. Explicit integration of positional and size
  information as Gaussian masks directs attention towards specific target
  regions. Experimental results on the nuScenes dataset demonstrate that incorporating
  object lists into vision-based methods significantly enhances detection accuracy.
  This validates the effectiveness of our model in bridging high-level and low-level
  fusion. Furthermore, our approach is adaptable and can be extended beyond
  detection. Future work will integrate velocity-aware DNF and temporal attention
  to improve velocity estimation and robustness to absent object lists through consistent
  temporal modeling. Additionally, we plan to investigate adaptive fusion strategies,
  where the model learns to weigh object list information based on reliability, reducing
  the impact of noisy detections and extend our fusion framework to multi-object
  tracking. Finally, comparison against traditional high-level fusion methods will
  be crucial to further assess its practical applicability in autonomous driving.

  % \section*{Acknowledgment}

  % The preferred spelling of the word ``acknowledgment'' in America is without
  % an ``e'' after the ``g''. Avoid the stilted expression ``one of us (R. B.
  % G.) thanks $\ldots$''. Instead, try ``R. B. G. thanks$\ldots$''. Put sponsor
  % acknowledgments in the unnumbered footnote on the first page.

  % \section*{References}

  % Please number citations consecutively within brackets \cite{b1}. The
  % sentence punctuation follows the bracket \cite{b2}. Refer simply to the reference
  % number, as in \cite{b3}---do not use ``Ref. \cite{b3}'' or ``reference \cite{b3}'' except at
  % the beginning of a sentence: ``Reference \cite{b3} was the first $\ldots$''

  % Number footnotes separately in superscripts. Place the actual footnote at
  % the bottom of the column in which it was cited. Do not put footnotes in the
  % abstract or reference list. Use letters for table footnotes.

  % Unless there are six authors or more give all authors' names; do not use
  % ``et al.''. Papers that have not been published, even if they have been
  % submitted for publication, should be cited as ``unpublished'' \cite{b4}. Papers
  % that have been accepted for publication should be cited as ``in press'' \cite{b5}.
  % Capitalize only the first word in a paper title, except for proper nouns and
  % element symbols.

  % For papers published in translation journals, please give the English
  % citation first, followed by the original foreign-language citation \cite{b6}.
  %%%%%%%%% REFERENCES
  {\small \bibliographystyle{IEEEtran} \bibliography{IEEEfull} }

@String(CVPR= {IEEE Conf. Comput. Vis. Pattern Recog.})

@String(CVPR  = {CVPR})

@ARTICLE{chong2000track,
  author={Chee-Yee Chong and Mori, S. and Barker, W.H. and Kuo-Chu Chang},
  journal={IEEE Aerospace and Electronic Systems Magazine}, 
  title={Architectures and algorithms for track association and fusion}, 
  year={2000},
  volume={15},
  number={1},
  pages={5-13},
  doi={10.1109/62.821657}}

@inproceedings{nuscenes2019,
  title={nuscenes: A multimodal dataset for autonomous driving},
  author={Caesar, Holger and Bankiti, Varun and Lang, Alex H and Vora, Sourabh and Liong, Venice Erin and Xu, Qiang and Krishnan, Anush and Pan, Yu and Baldan, Giancarlo and Beijbom, Oscar},
  booktitle={Proceedings of the IEEE/CVF conference on computer vision and pattern recognition},
  pages={11621--11631},
  year={2020}
}

@inproceedings{ambrosin2019design,
  title={Design of a misbehavior detection system for objects based shared perception V2X applications},
  author={Ambrosin, Moreno and Yang, Lily L and Liu, Xiruo and Sastry, Manoj R and Alvarez, Ignacio J},
  booktitle={2019 IEEE Intelligent Transportation Systems Conference (ITSC)},
  pages={1165--1172},
  year={2019},
  organization={IEEE}
}

@misc{mmdet3d2020,
    title={{MMDetection3D: OpenMMLab} next-generation platform for general {3D} object detection},
    author={MMDetection3D Contributors},
    howpublished = {\url{https://github.com/open-mmlab/mmdetection3d}},
    year={2020}
}

@inproceedings{liu2023petrv2,
  title={Petrv2: A unified framework for 3d perception from multi-camera images},
  author={Liu, Yingfei and Yan, Junjie and Jia, Fan and Li, Shuailin and Gao, Aqi and Wang, Tiancai and Zhang, Xiangyu},
  booktitle={Proceedings of the IEEE/CVF International Conference on Computer Vision},
  pages={3262--3272},
  year={2023}
}

@inproceedings{gao2021fast,
  title={Fast convergence of detr with spatially modulated co-attention},
  author={Gao, Peng and Zheng, Minghang and Wang, Xiaogang and Dai, Jifeng and Li, Hongsheng},
  booktitle={Proceedings of the IEEE/CVF international conference on computer vision},
  pages={3621--3630},
  year={2021}
}

@inproceedings{bai2022transfusion,
  title={Transfusion: Robust lidar-camera fusion for 3d object detection with transformers},
  author={Bai, Xuyang and Hu, Zeyu and Zhu, Xinge and Huang, Qingqiu and Chen, Yilun and Fu, Hongbo and Tai, Chiew-Lan},
  booktitle={Proceedings of the IEEE/CVF conference on computer vision and pattern recognition},
  pages={1090--1099},
  year={2022}
}

@inproceedings{zhou2019objects,
  title={Objects as Points},
  author={Zhou, Xingyi and Wang, Dequan and Kr{\"a}henb{\"u}hl, Philipp},
  booktitle={arXiv preprint arXiv:1904.07850},
  year={2019}
}

@inproceedings{li2022dn,
  title={Dn-detr: Accelerate detr training by introducing query denoising},
  author={Li, Feng and Zhang, Hao and Liu, Shilong and Guo, Jian and Ni, Lionel M and Zhang, Lei},
  booktitle={Proceedings of the IEEE/CVF Conference on Computer Vision and Pattern Recognition},
  pages={13619--13627},
  year={2022}
}

@InProceedings{Lee_2020_CVPR,
author = {Lee, Youngwan and Park, Jongyoul},
title = {CenterMask: Real-Time Anchor-Free Instance Segmentation},
booktitle = {Proceedings of the IEEE/CVF Conference on Computer Vision and Pattern Recognition (CVPR)},
month = {June},
year = {2020}
}

@article{MathWorksFusion,
  author = {Chong, C. Y. and Mori, S. and Barker, W. H. and Chang, K. C.},
  title = {Architectures and Algorithms for Track Association and Fusion},
  journal = {IEEE Aerospace and Electronic Systems Magazine},
  volume = {15},
  number = {1},
  year = {2000},
  pages = {5--13}
}

@article{castanedo2013fusion,
author = {Castanedo, Federico},
title = {A Review of Data Fusion Techniques},
journal = {The Scientific World Journal},
volume = {2013},
number = {1},
pages = {704504},
doi = {https://doi.org/10.1155/2013/704504},
url = {https://onlinelibrary.wiley.com/doi/abs/10.1155/2013/704504},
eprint = {https://onlinelibrary.wiley.com/doi/pdf/10.1155/2013/704504},
year = {2013}
}

@inproceedings{yan2023cross,
  title={Cross modal transformer: Towards fast and robust 3d object detection},
  author={Yan, Junjie and Liu, Yingfei and Sun, Jianjian and Jia, Fan and Li, Shuailin and Wang, Tiancai and Zhang, Xiangyu},
  booktitle={Proceedings of the IEEE/CVF international conference on computer vision},
  pages={18268--18278},
  year={2023}
}

@ARTICLE{chong2011fusion,
  author={Chong, Chee-Yee},
  journal={IEEE Control Systems Magazine}, 
  title={Tracking and Data Fusion: A Handbook of Algorithms (Bar-Shalom, Y. et al; 2011) [Bookshelf]}, 
  year={2012},
  volume={32},
  number={5},
  pages={114-116},
  doi={10.1109/MCS.2012.2204808}}

@ARTICLE{chen2003limits,
  author={Huimin Chen and Kirubarajan, T. and Bar-Shalom, Y.},
  journal={IEEE Transactions on Aerospace and Electronic Systems}, 
  title={Performance limits of track-to-track fusion versus centralized estimation: theory and application [sensor fusion]}, 
  year={2003},
  volume={39},
  number={2},
  pages={386-400},
  doi={10.1109/TAES.2003.1207252}}

@ARTICLE{thomopoulos1987optimal,
  author={Thomopoulos, Stelios C.A. and Viswanathan, Ramanarayanan and Bougoulias, Dimitrios C.},
  journal={IEEE Transactions on Aerospace and Electronic Systems}, 
  title={Optimal Decision Fusion in Multiple Sensor Systems}, 
  year={1987},
  volume={AES-23},
  number={5},
  pages={644-653},
  doi={10.1109/TAES.1987.310858}}

@INPROCEEDINGS{matzka2008multifusion,
  author={Matzka, Stephan and Altendorfer, Richard},
  booktitle={2008 IEEE International Conference on Multisensor Fusion and Integration for Intelligent Systems}, 
  title={A comparison of track-to-track fusion algorithms for automotive sensor fusion}, 
  year={2008},
  volume={},
  number={},
  pages={189-194},
  doi={10.1109/MFI.2008.4648063}}

@article{huang2022multi,
  title={Multi-modal sensor fusion for auto driving perception: A survey},
  author={Huang, Keli and Shi, Botian and Li, Xiang and Li, Xin and Huang, Siyuan and Li, Yikang},
  journal={arXiv preprint arXiv:2202.02703},
  year={2022}
}

@inproceedings{chen2023futr3d,
  title={Futr3d: A unified sensor fusion framework for 3d detection},
  author={Chen, Xuanyao and Zhang, Tianyuan and Wang, Yue and Wang, Yilun and Zhao, Hang},
  booktitle={proceedings of the IEEE/CVF conference on computer vision and pattern recognition},
  pages={172--181},
  year={2023}
}

@inproceedings{girshick2014rich,
  title={Rich feature hierarchies for accurate object detection and semantic segmentation},
  author={Girshick, Ross and Donahue, Jeff and Darrell, Trevor and Malik, Jitendra},
  booktitle={Proceedings of the IEEE conference on computer vision and pattern recognition},
  pages={580--587},
  year={2014}
}

@article{xu2023fusionrcnn,
  title={Fusionrcnn: Lidar-camera fusion for two-stage 3d object detection},
  author={Xu, Xinli and Dong, Shaocong and Xu, Tingfa and Ding, Lihe and Wang, Jie and Jiang, Peng and Song, Liqiang and Li, Jianan},
  journal={Remote Sensing},
  volume={15},
  number={7},
  pages={1839},
  year={2023},
  publisher={MDPI}
}

@inproceedings{chen2017multi,
  title={Multi-view 3d object detection network for autonomous driving},
  author={Chen, Xiaozhi and Ma, Huimin and Wan, Ji and Li, Bo and Xia, Tian},
  booktitle={Proceedings of the IEEE conference on Computer Vision and Pattern Recognition},
  pages={1907--1915},
  year={2017}
}

@inproceedings{ku2018joint,
  title={Joint 3d proposal generation and object detection from view aggregation},
  author={Ku, Jason and Mozifian, Melissa and Lee, Jungwook and Harakeh, Ali and Waslander, Steven L},
  booktitle={2018 IEEE/RSJ International Conference on Intelligent Robots and Systems (IROS)},
  pages={1--8},
  year={2018},
  organization={IEEE}
}

@article{spencer2004smart,
  title={Smart sensing technology: opportunities and challenges},
  author={Spencer Jr, BF and Ruiz-Sandoval, Manuel E and Kurata, Narito},
  journal={Structural Control and Health Monitoring},
  volume={11},
  number={4},
  pages={349--368},
  year={2004},
  publisher={Wiley Online Library}
}

@inproceedings{yin2021center,
  title={Center-based 3d object detection and tracking},
  author={Yin, Tianwei and Zhou, Xingyi and Krahenbuhl, Philipp},
  booktitle={Proceedings of the IEEE/CVF conference on computer vision and pattern recognition},
  pages={11784--11793},
  year={2021}
}

@inproceedings{lang2019pointpillars,
  title={Pointpillars: Fast encoders for object detection from point clouds},
  author={Lang, Alex H and Vora, Sourabh and Caesar, Holger and Zhou, Lubing and Yang, Jiong and Beijbom, Oscar},
  booktitle={Proceedings of the IEEE/CVF conference on computer vision and pattern recognition},
  pages={12697--12705},
  year={2019}
}
  % \begin{thebibliography}{00}
  % \bibitem{b1} G. Eason, B. Noble, and I. N. Sneddon, ``On certain integrals of Lipschitz-Hankel type involving products of Bessel functions,'' Phil. Trans. Roy. Soc. London, vol. A247, pp. 529--551, April 1955.
  % \bibitem{b2} J. Clerk Maxwell, A Treatise on Electricity and Magnetism, 3rd ed., vol. 2. Oxford: Clarendon, 1892, pp.68--73.
  % \bibitem{b3} I. S. Jacobs and C. P. Bean, ``Fine particles, thin films and exchange anisotropy,'' in Magnetism, vol. III, G. T. Rado and H. Suhl, Eds. New York: Academic, 1963, pp. 271--350.
  % \bibitem{b4} K. Elissa, ``Title of paper if known,'' unpublished.
  % \bibitem{b5} R. Nicole, ``Title of paper with only first word capitalized,'' J. Name Stand. Abbrev., in press.
  % \bibitem{b6} Y. Yorozu, M. Hirano, K. Oka, and Y. Tagawa, ``Electron spectroscopy studies on magneto-optical media and plastic substrate interface,'' IEEE Transl. J. Magn. Japan, vol. 2, pp. 740--741, August 1987 [Digests 9th Annual Conf. Magnetics Japan, p. 301, 1982].
  % \bibitem{b7} M. Young, The Technical Writer's Handbook. Mill Valley, CA: University Science, 1989.
  % \end{thebibliography}
  % \vspace{12pt}
  % \color{red}
  % IEEE conference templates contain guidance text for composing and formatting conference papers. Please ensure that all template text is removed from your conference paper prior to submission to the conference. Failure to remove the template text from your paper may result in your paper not being published.
\end{document}